\documentclass{article}
\pdfoutput=1

\usepackage{arxiv}

\usepackage[utf8]{inputenc} 
\usepackage[T1]{fontenc}    
\usepackage{hyperref}       
\usepackage{url}            
\usepackage{booktabs}       
\usepackage{amsfonts}       
\usepackage{nicefrac}       
\usepackage{microtype}      
\usepackage{lipsum}		
\usepackage{graphicx}
\usepackage{natbib}
\usepackage{doi}

\usepackage{etoolbox}
\patchcmd{\thebibliography}{\section*}{\section}{}{}

\title{Optimizing Neural Network for Computer Vision task in Edge Device }

\author{ \href{https://orcid.org/0000-0001-6214-4729}{\includegraphics[scale=0.06]{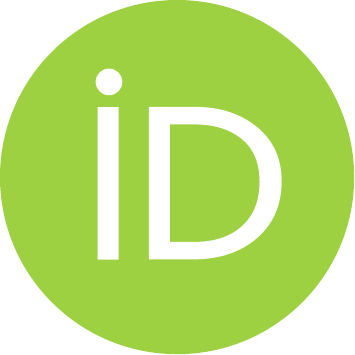}\hspace{1mm}Ranjith M S}\\
	\And
		Dr.~ S Parameshwara \\
		\And
	Pavan Yadav A \\
			\And
	Shriganesh Hegde \\
	\And
	Department of Electronics and Communication\\
	The National Institute of Engineering\\
}

\renewcommand{\headeright}{Research Paper}
\renewcommand{\undertitle}{Research Paper}

\hypersetup{
	pdftitle={Optimizing Neural Network for Computer Vision task in Edge Device},
	pdfsubject={q-bio.NC, q-bio.QM},
	pdfauthor={Ranjith M S},
	pdfkeywords={Embedded System, Edge Computing, Optimization, Deep learning, Computer Vision, Pruning, Quantization, ARM, convolution neural network},
}

\begin{document}
	\maketitle
	
	\begin{abstract}
		The field of computer vision has grown very rapidly in the past few years due to networks like convolution neural networks and their variants. The memory required to store the model and computational expense are very high for such a network limiting it to deploy on the edge device. Many times, applications rely on the cloud but that makes it hard for working in real-time due to round-trip delays. We overcome these problems by deploying the neural network on the edge device itself. The computational expense for edge devices is reduced by reducing the floating-point precision of the parameters in the model. After this the memory required for the model decreases and the speed of the computation increases where the performance of the model is least affected. This makes an edge device to predict from the neural network all by itself. 
	\end{abstract}


	\section{Introduction}
	With the rise of Artificial intelligence, it is realized that deep-learning-based approaches give satisfactory results compared to numerous other state-of-the-art schemes which are hand-engineered to all the computer vision tasks. The features learned from Convolutional neural networks outperform other hand-engineered feature-based methods like SIFT [\cite{r1}] and HoG [\cite{r2}] in computer vision tasks like image classification and object detection.\\
	The availability of large datasets and powerful computation devices made it possible to train the large and complex neural networks to obtain the desired performance on many computer vision tasks. The large amount of open-source pre-trained models trained on large datasets like ImageNet [\cite{r3}], MS-COCO [\cite{r4}], SHVN [\cite{r5}] created a large number of useful filters especially the features learned from the initial layer helps in transfer learning a lot. In transfer learning, most of the time only the last few layers of pre-trained models are modified and trained which counters the problem of having fewer data to a certain extent.\\ 
	Currently, a variety of embedded systems are deployed but the usage of neural networks is limited in edge devices like microcontrollers, Raspberry Pi. Household devices like Refrigerators, washing machines use a set of logic, rules for their automatic operations. By optimizing the network trained on a dataset traditional way of controlling can be replaced by intelligently monitoring the systems with the power of AI and neural networks [\cite{r6}]. Optimizing the convolutional neural network architectures like ResNet [~\cite{r7}], DenseNet [~\cite{r8}], AlexNet [~\cite{r9}] which are generally used in computer vision tasks allows creating many useful applications. for example, using gestures recognized from the neural network to perform a specific task based on the requirement, where the task and the gesture recognition are performed by the edge device itself. \\
	\subsection{Literature Survey for implementing and training the neural network}
	Since we use dense layers along with convolutional layers for our task we prefer He initialization [~\cite{r10}] as compared with Xavier Initialization ~\cite{r11} for initializing the weights while training the network. But using LSUV 
	initialization [~\cite{r12}] leads to the learning of very deep nets that produces networks with test accuracy better or equal to standard methods.\\
	“overfitting" is greatly reduced by randomly omitting half of the feature detectors on each training case. This prevents complex co-adaptations in which a feature detector is only helpful in the context of several other specific feature detectors. Instead, each neuron learns to detect a feature that is generally helpful for producing the correct answer given the combinatorially large variety of internal contexts in which it must operate.” [~\cite{r13}].\\
	The optimizers such as Adam [~\cite{r14}], RMS [~\cite{r15}] lead for faster convergence. And adaptive optimizers like LARS [~\cite{r16}], AdamW [~\cite{r17}] lead to better generalization.\\
	 “One of the key elements of super-convergence is training with one learning rate cycle and a large maximum learning rate” [~\cite{r18}].\\
	  To speed up training we use Batch Normalization [~\cite{r19}].\\
	  
	We evaluate the performance of the optimized model for recognizing hand gestures in raspberry pi. We collect the dataset for our task, in order to obtain a good dataset, we take images in different lightings, different cameras, and angles which helps in creating a robust good accurate model. The neural network is implemented and trained using a device like GPU which has strong computational power. Then the model is optimized by reducing the floating-point precision of trained parameters and by pruning the parameters. The model is then evaluated and deployed on edge device like raspberry pi.

	\section{Experimental Setup}
	\label{sec:Experimental Setup}
	\begin{figure}[h]
		
		{\includegraphics[width=14cm, height=7cm]{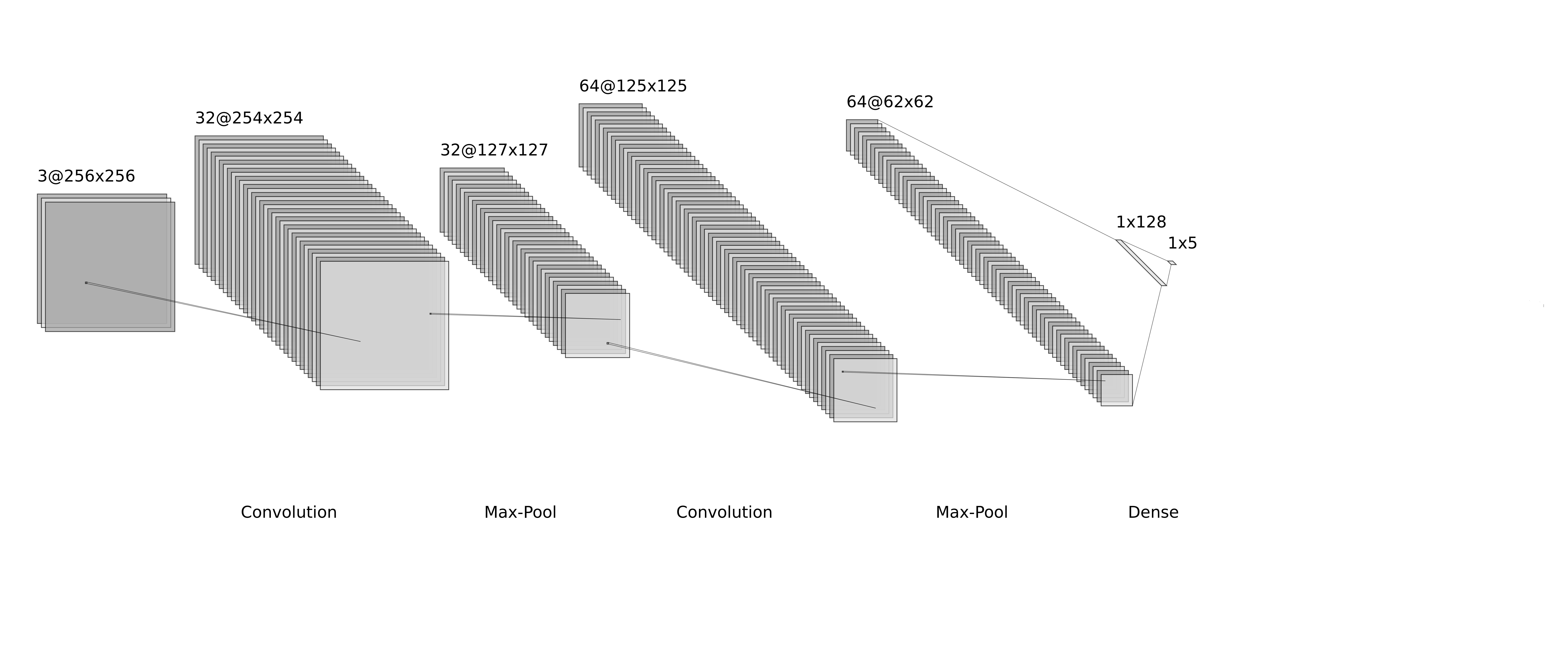}}
		
		\caption{Neural Network Architecture}
		\label{fig:fig1}
	\end{figure}
	
	\subsection{Preparation of dataset}
	Dataset is prepared by collecting images for each different class. i.e., for each unique gesture to be recognized. To obtain a robust model many numbers of images are captured in different light intensity, angles, background, distance, camera for every gesture to be recognized. Then images belonging to a unique class are stored in a separate folder. Different sized datasets of images are created i.e., captured images are scaled to 64×64, 96×96, 128×128, 256×256 hence each size dataset we will have the images for each class of that size. This is done to find the optimal size for input images such that the images with less resolution are trained to the desired accuracy by which we can reduce the model size and hence computations required for prediction.  \\
	\subsubsection{Steps}
	\begin{itemize}
		\item Collect the images for each unique gesture under different lighting conditions, background, camera, angle and save each gesture image in their respective folders.
		\item Create a unique folder for all the required sizes of images (E.g., folder\_64, folder\_96 …etc)
		\item Under each different folder (folders created for the required size dataset) create the folders for each unique class (gesture) (E.g., one, two, three …. etc). 
		\item Read the images from the original folder resize them and save accordingly. We have used OpenCV to read images, resize them and store them accordingly. E.g. read an image of class ‘one’
		\item resize it to 64×64, 96×96, 128×128, 256×256 and store them in folder ‘one of ‘folder\_64’, ‘folder\_96’, ‘folder\_128’, ‘folder\_256’ respectively
	\end{itemize}
\begin{figure}[h]
	\centering
	
	{\includegraphics[width=15cm, height=7cm]{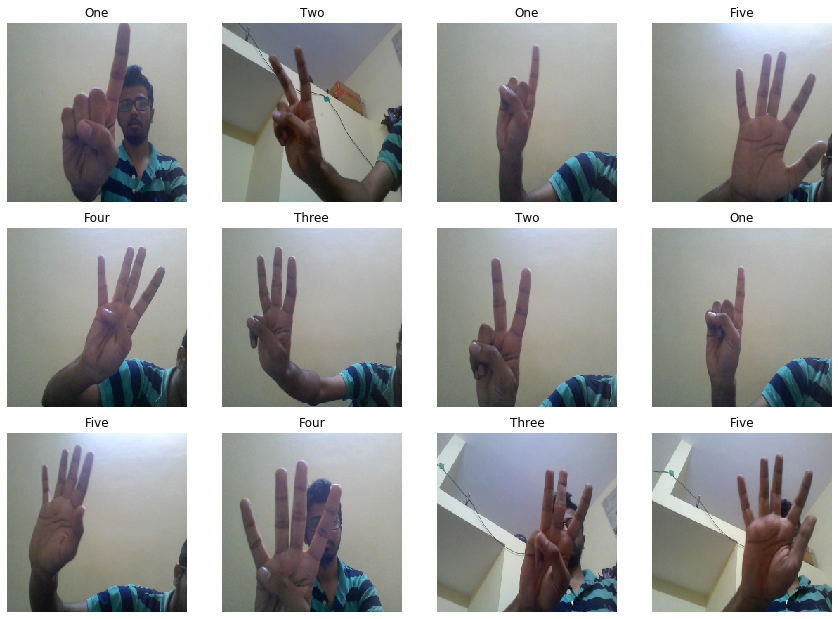}}
	
	\caption{Sample images in dataset}
	\label{fig:fig01}
\end{figure}

	\subsection{Training the Network}
	TensorFlow [~\cite{r20}] is used to create the network and load the dataset during training. The images are normalized during every iteration before feeding into the network and a few augmentation’s like horizontal flips were performed in order to generalize the model well. Dataset is split into 85\% for training and 15\% for validation. Fig-2 shows the network architecture for the training image dataset of size 256×256 with 3 channels (RGB). The normalized input images are 2D-convolved with a kernel size of 3×3 and 32 such filters are used, and the output is passed through the non-Linearity activation function ReLU [21]. Then the 2D-max-pooling (kernel size 2×2) is applied followed by Dropout. Then the output is 2D-convolved with a kernel size of 3×3 (64 filters) and passing through ReLU activation followed by 2D-max-pooling and Dropout. Then the output is flattened followed by Dropout and 2 Dense layers having 128 and 5 (as the number of unique gestures is 5) hidden units respectively. The first hidden layer has the ReLU activation function and the latter has SoftMax activation.

\begin{table}[h]
	\caption{Network structure}
		\centering
	\begin{tabular}{lll}
		\toprule
		Layer         & Output shape         & No. of parameters \\
		\toprule
		Input         & (None, 256,256,3)    & 0                 \\
		Conv2D        & (None, 254,254,32)   & 896               \\
		Max-pooling2D & (None, 127,127,32)   & 0                 \\
		Dropout       & (None,   127,127,32) & 0                 \\
		Conv2D        & (None,   125,125,64) & 18496             \\
		Max-pooling2D & (None, 62,62,64)     & 0                 \\
		Dropout       & (None, 62,62,64)     & 0                 \\
		Flatten       & (None,246016)        & 0                 \\
		Dropout       & (None,246016)        & 0                 \\
		Dense         & (None,128)           & 31490176          \\
		Dense         & (None,5)             & 645          \\
		\bottomrule  
		Total params & 31510213 \\
		  \bottomrule
	\end{tabular}

\end{table}
The model is trained using Tesla-K80 GPU which can do up to 8.73 Teraflops single-precision and up to 2.91 Teraflops double-precision performance with NVIDIA GPU Boost. The model is set to train for a maximum of 25 epochs provided the validation loss will not increase continuously for 3 epochs in that case the training is stopped, and the model weights are saved. \\
The model is trained using Adam optimizer with a batch size of 32. We used Categorical cross-entropy as the loss function to train the model. categorical cross-entropy is given by,
\begin{equation}
	C E=-\sum_{i}^{c} t_{i} \log (f(s) i)
\end{equation}
Where f(s)i represents SoftMax function and is given by,
\begin{equation}
	f(s){i}=\frac{e^{S{i}}}{\sum_{j}^{C} e^{s_{j}}}
\end{equation}
Where C represents the total number of class and sj are the score inferred by the neural network for each class in C.\\

\begin{figure}[!htbp]
	\begin{minipage}[t]{0.45\linewidth}
		\includegraphics[width=\linewidth]{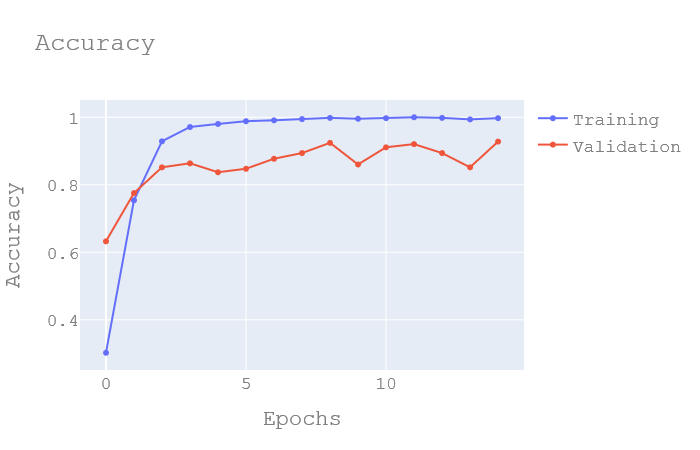}
		\caption{variation of accuracy for 256×256}
		\label{fig4}
	\end{minipage}%
	\hfill%
	\begin{minipage}[t]{0.45\linewidth}
		\includegraphics[width=\linewidth]{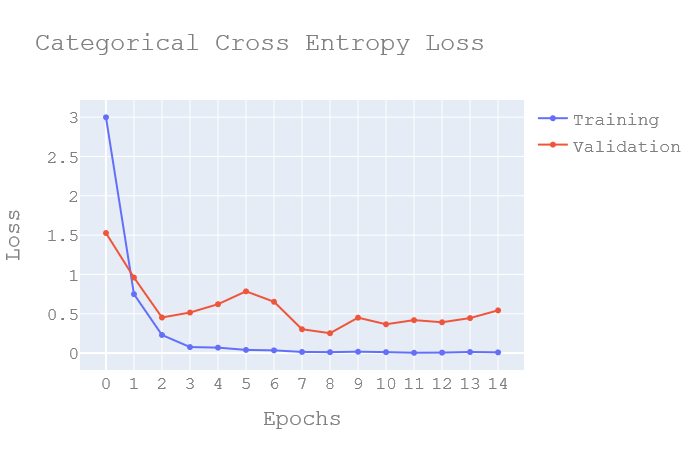}
		\caption{variation of loss for 256×256}
		\label{fig3}
	\end{minipage} 
\end{figure}

Figure~\ref{fig3} indicates how training and validation loss varied during training in each epoch and we see that both training and validation loss reach a very low value by 14 epochs. And the training is stopped after 14 epochs as the validation loss keeps increasing by a small value. The best weights are saved.

Figure ~\ref{fig4} indicates how accuracy varied during training after each epoch. Training and validation accuracy reach desired values by the end of 14 epochs.

\begin{figure}[!htbp]
	\begin{minipage}[t]{0.45\linewidth}
		\includegraphics[width=\linewidth]{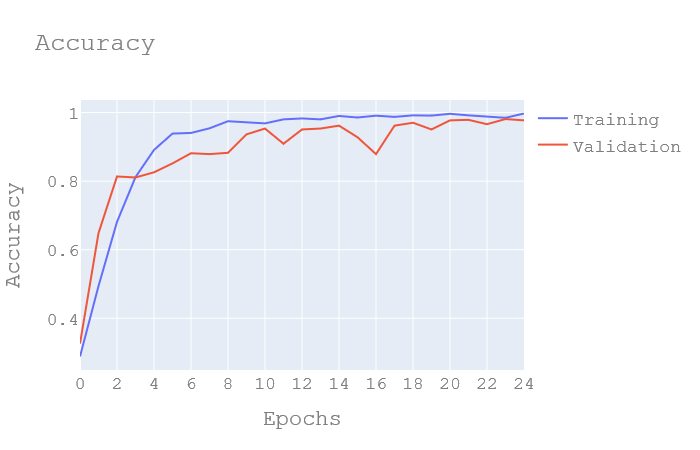}
		\caption{variation of accuracy for 64×64}
		\label{f5}
	\end{minipage}%
	\hfill%
	\begin{minipage}[t]{0.45\linewidth}
		\includegraphics[width=\linewidth]{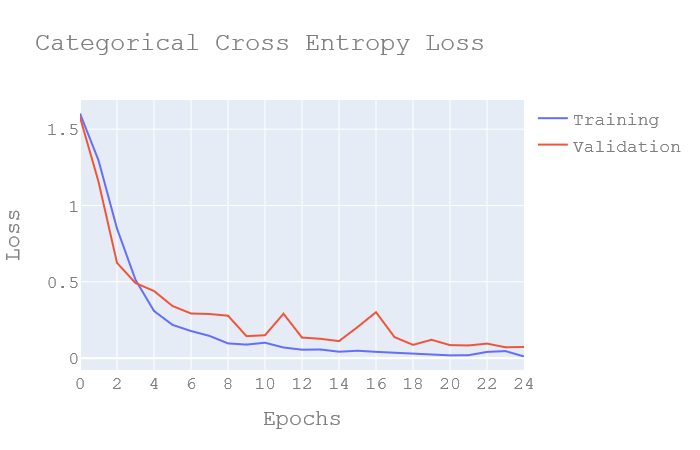}
		\caption{variation of loss for 64×64}
		\label{f6}
	\end{minipage} 
\end{figure}

\begin{figure}[!htbp]
	\begin{minipage}[t]{0.45\linewidth}
		\includegraphics[width=\linewidth]{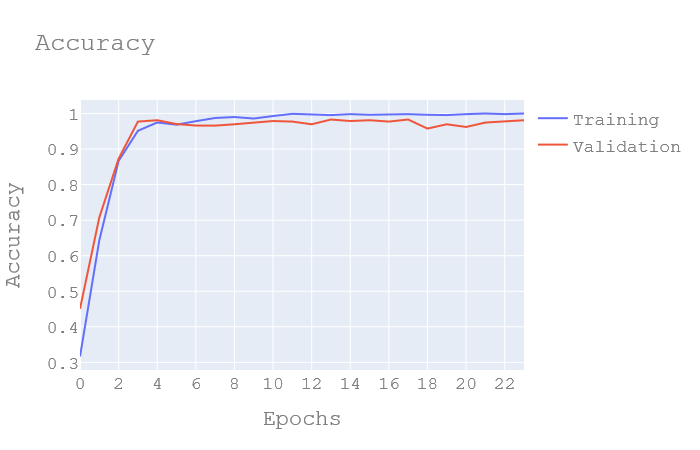}
		\caption{variation of accuracy for 96×96}
		\label{f7}
	\end{minipage}%
	\hfill%
	\begin{minipage}[t]{0.45\linewidth}
		\includegraphics[width=\linewidth]{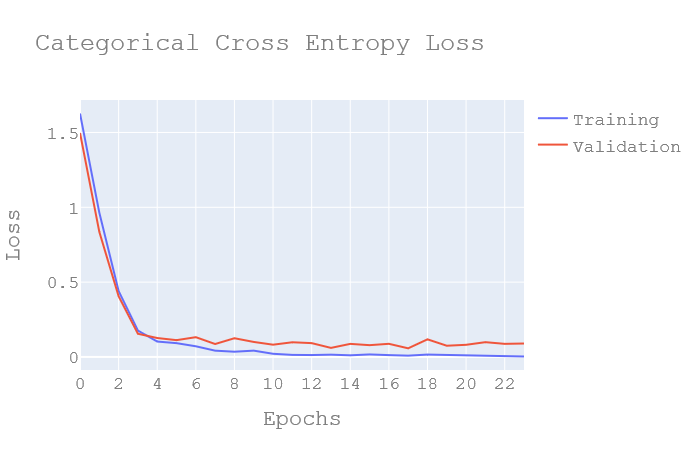}
		\caption{variation of loss for 96×96}
		\label{f8}
	\end{minipage} 
\end{figure}

\begin{figure}[!htbp]
	\begin{minipage}[t]{0.45\linewidth}
		\includegraphics[width=\linewidth]{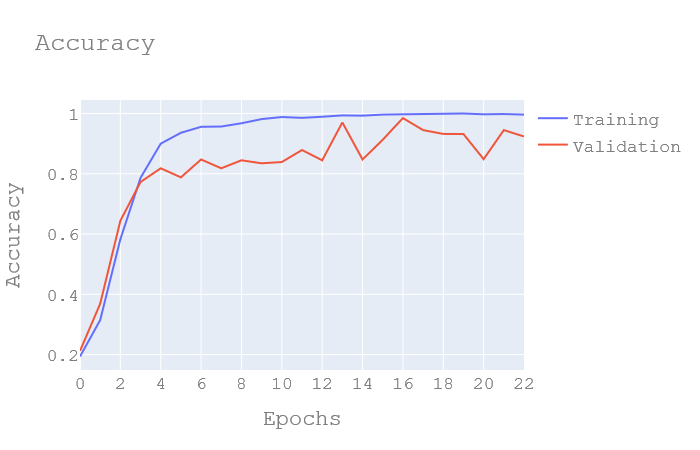}
		\caption{variation of accuracy for 128×128}
		\label{f10}
	\end{minipage}%
	\hfill%
	\begin{minipage}[t]{0.45\linewidth}
		\includegraphics[width=\linewidth]{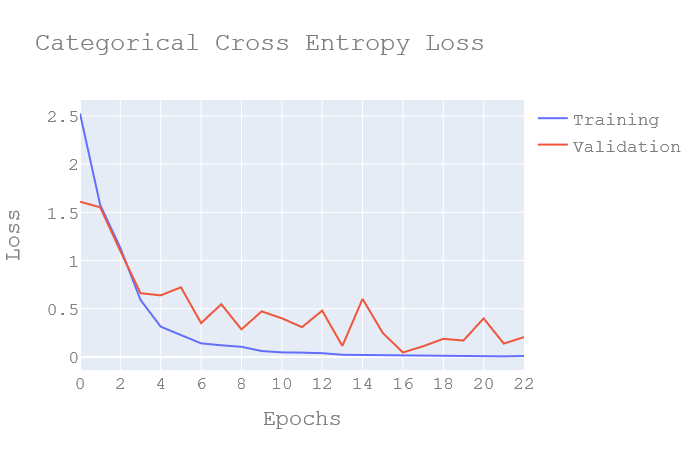}
		\caption{variation of loss for 128×128}
		\label{f9}
	\end{minipage} 
\end{figure}

Similar results were obtained when tried with different image sizes and found that image size with 96x96 works best among all during the test time. Figures ~\ref{f5} - ~\ref{f9} indicates the variation of accuracy and loss for different input sizes of the images during training.  Since our aim is to fit the model in edge device memory, it is suitable to choose the model which gives desirable accuracy keeping image size as low as possible. The model trained for image size with 96 × 96 was obtained over 98\% accuracy on the validation set indicating the best input size. The model trained with 128×128 and 256×256 images gave very good training accuracy (nearly 100\%) but were able to give only around 92\% accuracy on the validation set indicating high variance of the model. This may be due to fewer data and the usage of large neural networks. 
\begin{figure}[h]
	\centering
	
	{\includegraphics[width=15cm, height=5cm]{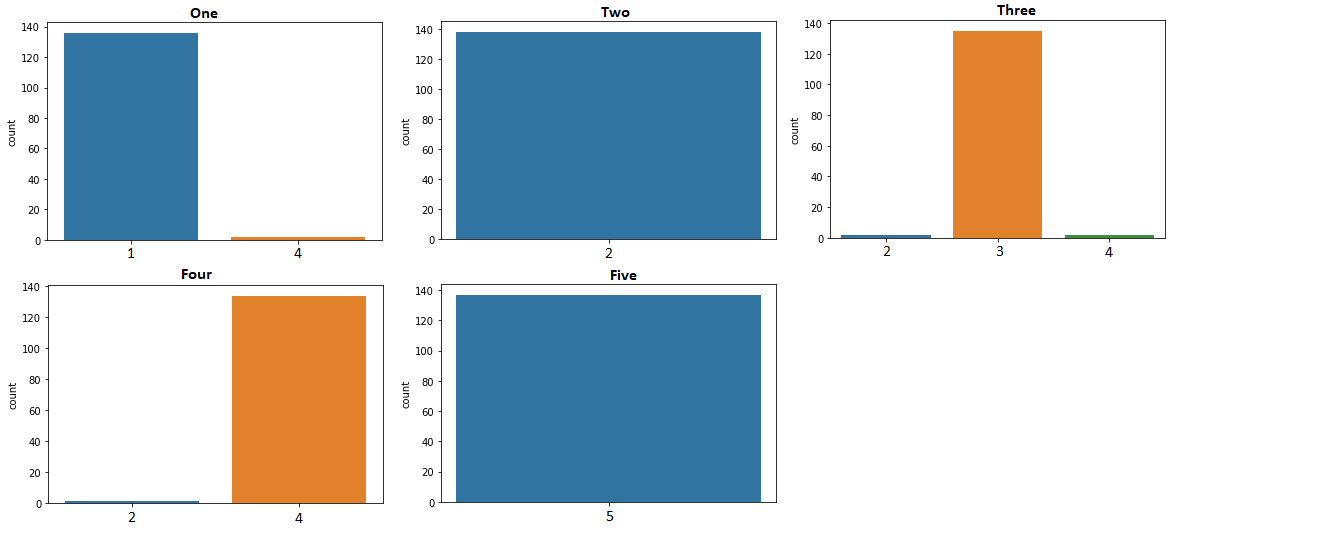}}
	
	\caption{Misclassifications of the network}
	\label{f11}
\end{figure}

In many a case one gesture will be similar to the others which usually makes models to perform badly on them. To reduce such misclassifications higher weight is assigned for the loss of such classes. Here slightly higher weight is applied to gestures ‘one’, ‘three’, ‘four’ hence we see a smaller number of misclassifications (Figure- ~\ref{f11}). Else the misclassification for those gesture will be slightly high.
	\section{Optimization}
	\label{sec:optimization}
	The trained model able to perform classifications has the parameters with the floating-point precision of 64-bit occupies more memory while performing the inference. Since floating-point arithmetic operations need more machine cycles and the machine cycles required for an operation increases with increase in floating-point precision. By reducing the floating-point precision machine cycles required to perform the operation reduces along with the memory. The reduction in the floating-point precision does not affect the accuracy more and in rare cases, certain models may gain some accuracy as a result of the optimization process . Since deep neural networks require more floating-point operations for its prediction reducing floating-point precision makes it suitable for edge devices.\\
	Pruning of parameters is done properly using the representative dataset during the model optimization i.e., reduce parameter count by structured pruning. We use TensorFlow lite for reducing the floating-point precision of our trained TensorFlow model which leads to optimization. \\
	
	\begin{figure}[h]
		\centering
		
		{\includegraphics[width=12cm, height=6cm]{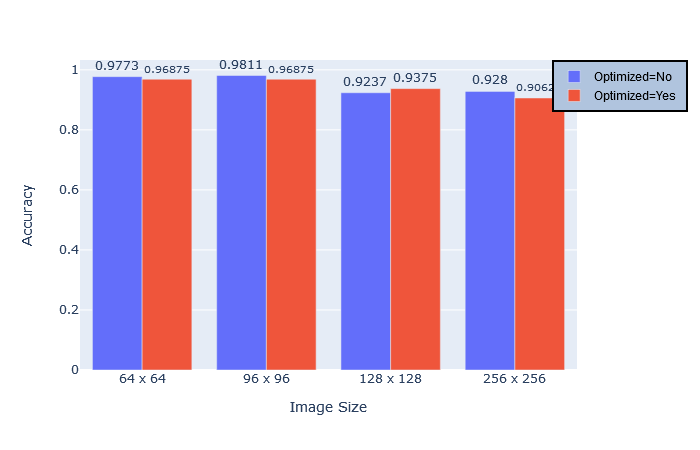}}
		
		\caption{Variation of accuracy with optimization}
		\label{f12}
	\end{figure}
	Figure- ~\ref{f12} indicates the variation of accuracy with optimization for models trained on different image sizes. For a non-optimized models accuracy indicates the accuracy on validation set (100 examples) and for optimized models’ accuracy is calculated on test set of 32 examples. We observe that the model optimized for the Neural network trained on 96×96 images has good accuracy and performance during inference. Since all the models have similar accuracy, with models being trained on 96×96 and 64×64 images are slightly better.
	
		\begin{figure}[h]
		\centering
		
		{\includegraphics[width=12cm, height=6cm]{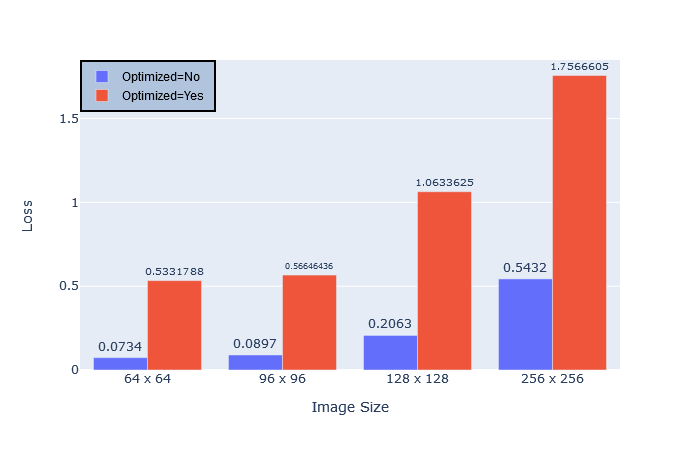}}
		
		\caption{Variation of loss with optimization}
		\label{f13}
	\end{figure}

Although we have similar accuracy between optimized and non-optimized models the loss of optimized model is comparatively high. Figure ~\ref{f13} shows the variation of loss with optimization for neural networks trained on different image size. The loss for non-optimized models is computed on validation set of 100 examples and loss for optimized models is computed on test set of 32 examples.

	\begin{figure}[h]
	\centering
	
	{\includegraphics[width=12cm, height=6cm]{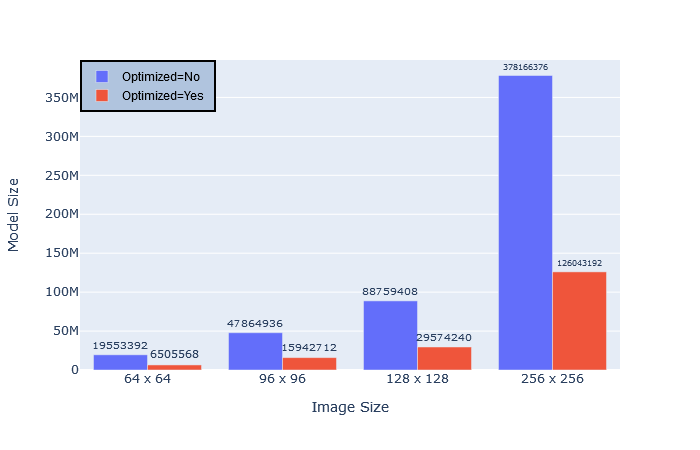}}
	
	\caption{Variation of model size with optimization.}
	\label{f14}
\end{figure}
Variation of model size with optimization for different input image sizes is shown in the Figure ~\ref{f14}. The size of the model reduces nearly to 33\% of initial size after optimization. We observe that the model with input image shapes of 96×96 and 64×64 after optimization reduces to 15.9MB and 6.5MB respectively which could be deployed in edge devices.

\begin{table}[h]
	\caption{Network structure}
	\centering
	\label{t2}
	\begin{tabular}{lllll}
		\toprule
		Image   Size & Optimized & Loss     & Accuracy & Model   Size \\
		\toprule
		128 x 128    & No        & 0.206300 & 0.92370  & 88759408     \\ \toprule
		128 x 128    & Yes       & 1.063362 & 0.93750  & 29574240     \\ \toprule
		64 x 64      & No        & 0.073400 & 0.97730  & 19553392     \\ \toprule
		64 x 64      & Yes       & 0.533179 & 0.96875  & 6505568      \\ \toprule
		96 x 96      & No        & 0.089700 & 0.98110  & 47864936     \\ \toprule
		96 x 96      & Yes       & 0.566464 & 0.96875  & 15942712     \\ \toprule
		256 x 256    & No        & 0.543200 & 0.92800  & 378166376    \\ \toprule
		256 x 256    & Yes       & 1.756660 & 0.90625  & 126043192   \\
		\bottomrule
	\end{tabular}
\end{table}
Table- ~\ref{t2} shows the accuracy and loss of the model before and after the optimization and the respective model size in bytes. Loss and accuracy for the optimized model is computed on the test set containing 32 examples and for non-optimized models they are computed on validation set of 100 examples.

\section{Conclusion}
The optimized models for this use case could be easily deployed on edge device like raspberry pi. Since the memory required for the models with input image shape 96×96 and 64×64 is few MB it could be fulfilled by edge devices easily. In the cases if the edge computing is hard then the aid from neural compute stick could be obtained for the deployment of the model.

{
\small
\nocite{*}
\newcounter{bibcount}
\bibliographystyle{ieee_fullname}

}  


	
	
	

\end{document}